\documentclass{article}

\usepackage[preprint]{neurips_2024}

\usepackage[utf8]{inputenc} 
\usepackage[T1]{fontenc}    
\usepackage{hyperref}       
\usepackage{url}            
\usepackage{booktabs}       
\usepackage{amsfonts}       
\usepackage{nicefrac}       
\usepackage{microtype}      
\usepackage{xcolor}         

\usepackage{amsmath}
\usepackage{graphicx}

\usepackage{ifthen}
\newboolean{includeBackgroundOnHaas}
\setboolean{includeBackgroundOnHaas}{false} 

\title{The impact of conformer quality on learned representations of molecular conformer ensembles}

\author{%
  Keir Adams \ \ \ \ \ \ \ \ \ Connor W. Coley\\ 
  \\
  Department of Chemical Engineering\\
  Massachusetts Institute of Technology\\
  \texttt{\{keir, ccoley\}@mit.edu} \\
}

\begin{document}

\maketitle

\begin{abstract}
Training machine learning models to predict properties of molecular conformer ensembles is an increasingly popular strategy to accelerate the conformational analysis of drug-like small molecules, reactive organic substrates, and homogeneous catalysts. For high-throughput analyses especially, trained surrogate models can help circumvent traditional approaches to conformational analysis that rely on expensive conformer searches and geometry optimizations. 
Here, we question how the performance of surrogate models for predicting 3D conformer-dependent properties (of a single, active conformer) is affected by the quality of the 3D conformers used as their input. How well do lower-quality conformers inform the prediction of properties of higher-quality conformers? Does the fidelity of geometry optimization matter when encoding random conformers? For models that encode \emph{sets} of conformers, how does the presence of the ``active'' conformer that induces the target property affect model accuracy? How do predictions from a surrogate model compare to estimating the properties from cheap ensembles themselves? We explore these questions in the context of predicting Sterimol parameters of conformer ensembles optimized with density functional theory. Although answers will be case-specific, our analyses provide a valuable perspective on 3D representation learning models and raise practical considerations regarding when conformer quality matters.

\end{abstract}

\section{Introduction and Background}
\label{introduction}

Modeling properties of molecular conformer ensembles is ubiquitous across computational chemistry \citep{howard1988analysis, liu2022auto3d, davies2012conformational, hawkins2017conformation}, data-driven homogeneous catalyst design \citep{guan2018aaron, rosales2019rapid, hansen2016prediction, gensch2022comprehensive,brethome2019conformational}, and \textit{in silico} drug discovery\citep{gilson2007calculation, meng2011molecular, kuntz1982geometric, perola2004conformational}. For instance, identifying the lowest-energy conformer of a flexible molecule and simulating its electronic properties is standard practice in quantitative-structure-activity-relationship (QSAR) modeling of molecular function. It is also common to consider the contributions of other low-lying conformational states by computing Boltzmann-averaged properties of conformer ensembles \citep{mezei1986free, guan2018aaron, haas2022predicting}, which may better recapitulate experimental observables. Other studies simulate conformer ensembles as a means to identify an ``active'' conformer that is not necessarily low in energy, but is responsible for inducing activity. For example, identifying the binding pose of a ligand in protein-ligand docking or finding the most reactive transition state geometry of a stabilizing catalyst are common tasks in molecular design \citep{forli2016computational, friesner2004glide, guan2018aaron, jacobson2017automated,besora2011importance}. 

Simulating conformer ensembles typically follows a routine workflow \citep{friedrich2017benchmarking, hawkins2017conformation}. Given a molecular graph, 
an initial conformer search is performed to enumerate candidate conformers, usually with inexpensive simulation tools like molecular mechanics \citep{mohamadi1990macromodel}, stochastic embedding algorithms based on distance geometry \citep{riniker2015better}, or torsional scans \citep{hawkins2010conformer}. This conformer ensemble is then downsampled via clustering \citep{kim2017comparison, yongye2010dynamic} and/or energy filtering to obtain a reasonably-sized set of representative conformers that approximate the distribution of conformers that are thermodynamically accessible to the molecule under experimental conditions. These selected conformers may then undergo further geometry optimization at higher levels of theory (e.g., Density Functional Theory; DFT) to yield a final set of higher-quality conformers, which are then used for property estimations.

For properties that are sensitive to molecular structure, exhaustive conformer enumeration and rigorous geometry optimization are often needed for agreement with experiments \citep{laplaza2024overcoming, guan2018aaron}. 
However, high-throughput virtual screening often employs computational shortcuts to make analyses tractable. Restricting the conformer search, pruning ensembles, or using cheaper geometry optimizations are common tactics to accelerate conformational analysis. The consequences of such shortcuts are case-specific: Are the downstream properties of interest sensitive to geometric quality? Will missing certain conformations yield qualitatively inaccurate downstream predictions? Often, properties derived from conformer ensembles are used as input features for surrogate models that predict molecular activity. Do other sources of epistemic uncertainty, or the aleatoric uncertainty of the regression target, outweigh error caused by using imprecise ensembles?

The emergence of machine learning (ML)-based surrogate property prediction models adds further nuance to these questions of when, and how much, conformer quality matters. To reduce costs of property estimation, many studies train machine learning models to predict conformer-level properties given the conformer's geometry as input to the model, such as when predicting NMR shieldings from DFT-optimized geometries \citep{guan2021real,unzueta2021predicting}. This can be especially useful for high-throughput studies, where numerous property estimations are needed. 
Although this strategy may reduce the cost of property evaluation, conformer search and geometry optimization still present a bottleneck, particularly when multiple conformers are evaluated. Instead, recent studies have considered training ML surrogate models to \textit{predict} the properties of high-quality conformer ensembles using \textit{lower-quality geometries} or other cheap molecular featurizations as model inputs \citep{gensch2022comprehensive, haas2024rapid}. 
This is notably distinct from the conventional property prediction setting where the property is of a specific conformer that the model directly encodes, like with neural network potentials. In this setting, the predicted properties are of conformer ensembles that the model does \textit{not} have access to. 
As an illustrative example, \citet{haas2024rapid} trained surrogate models operating on cheap MMFF94-level conformers to rapidly predict over 80 steric, electronic, and stereoelectronic descriptors of DFT-optimized conformer ensembles of carboxylic acids and amines\ifthenelse{\boolean{includeBackgroundOnHaas}}{ (App. \ref{app:haas}).}{.}
These cheap surrogate models enabled rapid descriptor estimation for large chemical libraries that would be burdensome to obtain via traditional conformational analysis. 

\begin{figure}[t]
  \begin{center}
    \includegraphics[width=1.0\textwidth]{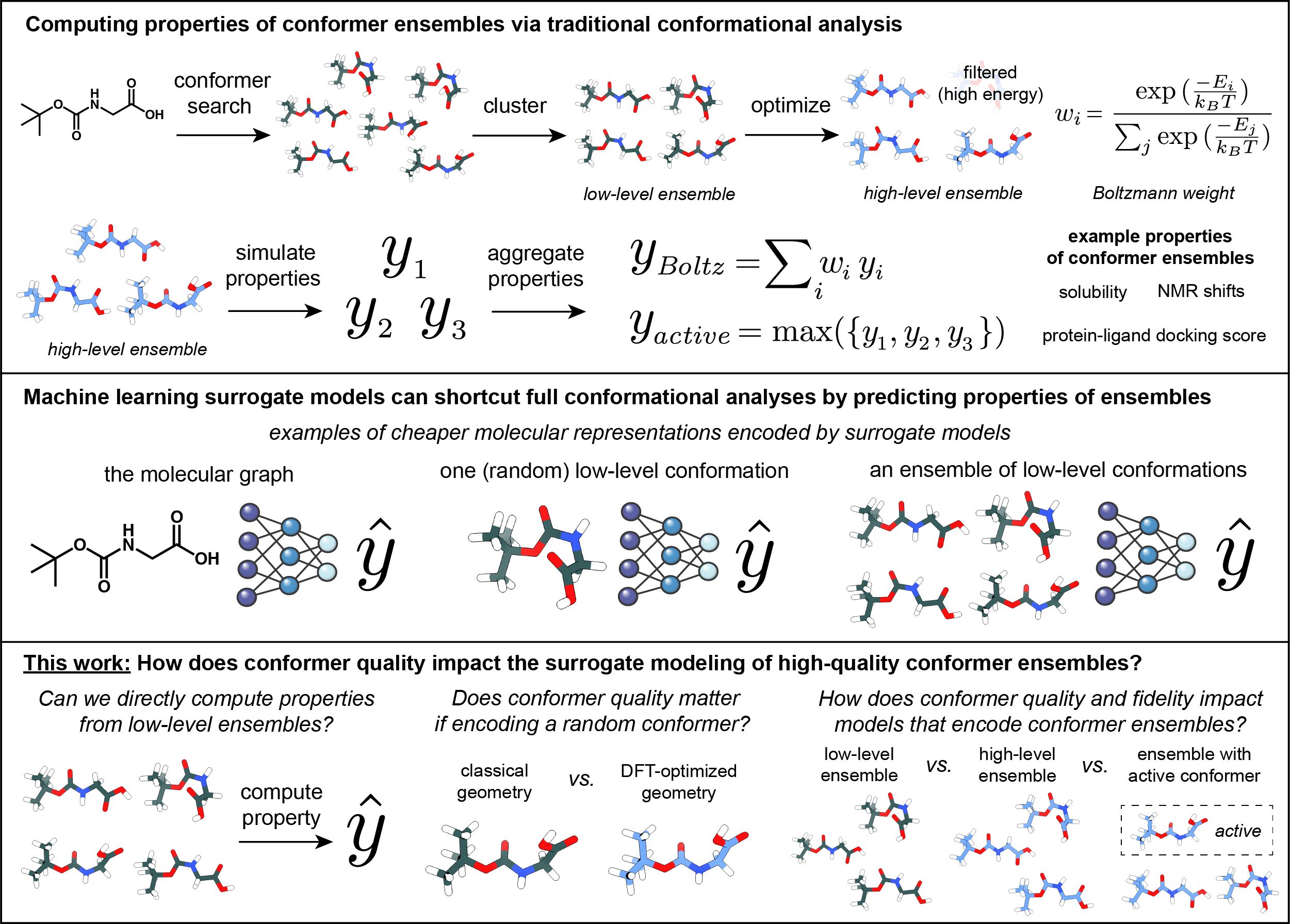}
  \end{center}
  \caption{(Top) Simulating properties of conformer ensembles typically involves a multi-step workflow beginning with the generation of high-quality conformer ensembles, which requires an initial conformer search, structural clustering, (high-level) geometry optimization, and energy filtering. Properties are then simulated for each conformer individually, followed by some kind of aggregation (e.g., Boltzmann-averaging, or using the maximum simulated property value of an ``active'' conformation). (Middle) Once trained, machine learning models can be used to shortcut expensive conformational analyses by directly predicting the ensemble-level properties from cheap-to-simulate molecular representations like a molecular graph, a single low-level conformation, or a set of low-level conformations. (Bottom) In this work, we consider how conformer quality impacts the performance of machine learning models that are trained to predict properties of high-quality conformer ensembles.}
  \label{fig:overview}
\vspace{-10pt}
\end{figure}

Using ML surrogate models to predict properties of high-quality conformer ensembles from less precise molecular representations presents subtle representation learning challenges that are as yet underexplored by the ML community. A straightforward approach is to predict the ensemble-level properties with standard graph neural networks (GNNs) that encode just the (2D) molecular graph. Although such GNNs are not 3D-aware, this strategy is arguably well-motivated: the conformer ensemble of a molecule is implicitly a function of its (stereo-specific) molecular graph, and hence GNNs \textit{should} be able to learn properties of conformer ensembles even while neglecting explicit 3D structural information. In practice, however, this approach can empirically underperform due to lacking 3D awareness \citep{haas2024rapid, van20243dreact}. 
Alternatively, 3D-aware ML models may be used to encode either a \textit{representative} conformation from the ensemble (i.e., an approximation of the lowest-energy conformer or other ``active" conformer), a \textit{random} conformation sampled from an ensemble of plausible low-energy conformers without necessarily being the \emph{lowest}-energy itself, or a full \textit{set} of conformations optimized at either low or high levels of theory. Each of these approaches has drawbacks. Whereas a \textit{representative} conformer may be relatively information-rich compared to a \textit{random} conformer, identifying such informative conformations may be computationally expensive, defeating the purpose of using \textit{cheap} molecular representations as model inputs. However, a \textit{random} conformer -- even if optimized at a high level of theory -- may be too noisy for the 3D model to effectively learn anything from its 3D geometry. Finally, whereas one might expect that encoding a \textit{set} of conformers would be advantageous for an ML model that predicts properties of conformer ensembles, empirical studies have found mixed (often lackluster) results with this strategy \citep{axelrod2023molecular,zhu2023learning,liu2021fast}, typically at great computational expense. In all three cases, it is also unclear how the geometric quality of the encoded conformer(s) impacts representation learning capabilities.

In this work, we explore three questions related to how  conformer quality impacts the prediction of properties of high-quality conformer ensembles with 3D machine learning surrogate models:
\begin{enumerate}
    \item What are the trade-offs between using ML surrogate models to \textit{predict} properties of high-quality ensembles 
    versus just computing the properties from cheap conformer ensembles? This question applies to settings where the property is relatively inexpensive to derive from a given conformer compared to the cost of conformer generation itself. 

    \item When encoding a single \textit{random} conformer to predict properties of high-quality conformer ensembles, does the \textit{local} geometric quality of the encoded conformer matter? Here, ``local geometric quality" refers to the relative fidelity of geometry optimization (e.g., DFT > xTB > MMFF94 > ETKDG).

    \item How does the local geometric quality and global structural fidelity of an encoded \textit{set} of conformers affect the ability of a 3D ML surrogate model to predict properties of high-quality conformer ensembles? In contrast to local geometric quality, ``global structural fidelity" refers to whether the encoded set contains a conformer that exactly matches or closely approximates the ground-truth ``active'' conformation that induces the target property. 
\end{enumerate}

We study these questions by making use of the datasets introduced by \citet{haas2024rapid}, but focus our analysis on certain steric parameters (Sterimol-L and Sterimol-B5) of carboxylic acids that can be quickly computed from a molecular conformation optimized at any level of theory. We specifically consider the minimum and maximum values of these two Sterimol parameters among the conformers in each ensemble, which can be considered to be induced by certain ``active'' conformers -- i.e., the conformers with the minimum/maximum Sterimol parameter. Importantly, we are most interested in settings where this ``active'' conformer is not known or even seen by the model. 
Whereas \citet{haas2024rapid} trained 2D and 3D GNNs (encoding \textit{random} MMFF94 conformers) to predict these ensemble-level properties\ifthenelse{\boolean{includeBackgroundOnHaas}}{ (see App. \ref{app:haas} for details),}{,}
here we consider the impact of training on higher-quality geometries such as xTB-optimized geometries and the DFT-optimized geometries used to derive the ground truth property labels themselves. We further analyze models which encode \textit{sets} of 3D conformers optimized at various levels of theory in order to investigate others' observations that encoding multiple conformers at once rarely improves performance.
Although we specifically analyze Sterimol descriptor prediction in this work as a tractable case study, our general questions and analyses aim to provide a new perspective on the representation learning capabilities of 3D surrogate machine learning models while prompting further discussion on when conformer quality matters.

\section{Related Work}
\label{relatedwork}

\textbf{Alternative machine learning strategies to accelerate conformational analysis of small molecules.}
In this work, we analyze models that seek to shortcut expensive conformational analyses by predicting properties of expensive-to-simulate conformer ensembles from cheaper molecular representations. We emphasize that this approach is not the only ML strategy that has been proposed to accelerate or otherwise circumvent traditional analyses. Most notable are works that train neural network potentials (NNPs) to serve as a drop-in replacement for DFT-based energy/force evaluations, thereby permitting faster geometry optimizations \citep{smith2017ani, zubatyuk2019accurate, anstine2024aimnet2}. Auto3D \citep{liu2022auto3d}, for instance, employs NNPs to rapidly optimize conformer ensembles at the $\omega$B97x/6-31G$^*$ level of theory. However, such NNPs do not natively replace the initial conformer search. To this end, many ML-based conformer generators have been proposed to accelerate or improve the coverage of traditional conformer generation \citep{ganea2021geomol, jing2022torsional}, particularly for small drug-like molecules. Finally, varied studies train ML surrogate models to replace DFT-based property calculations (e.g., given a known geometry) \citep{reiser2022graph}, which is most helpful if computing properties from a given geometry is relatively expensive compared to obtaining the geometry itself. In principle, combining all three strategies into one end-to-end workflow could vastly accelerate traditional conformational analysis without needing shortcuts.

\textbf{Predicting properties of conformer ensembles from cheaper 3D molecular representations.}
Multiple studies have considered training ML surrogate models to predict properties of conformer ensembles from encodings of cheap 3D molecular structures. In general, prior work can be categorized based on (1) whether the ground-truth property labels are experimentally derived (and hence an implicit function of the unknown experimental conformer ensemble), or are derived from explicitly simulated conformer ensembles (our case); (2) whether the models encode single conformer instances or sets of conformers via multi-instance learning \citep{dietterich1997solving, maron1997framework, ilse2018attention, zankov2024chemical}; and (3) whether the encoded conformers are of the same geometric quality as the ground truth simulated ensembles. For instance, \citet{axelrod2023molecular, zahrt2019prediction, zankov2021qsar, zankov2023multi, weinreich2021machine} all train ML models to predict experimental observables (e.g., ligand bioactivity, catalyst selectivity, or solvation free energies) from sets of simulated conformers. In contrast, \citet{chuang2020attention} and \citet{zhu2023learning} train ML-based set-encoders to predict properties of simulated conformer ensembles. Both studies primarily consider models that encode the same conformer ensembles (or subsets thereof) as the ensembles from which the target properties were originally derived. \citet{cremer2023equivariant} and \citet{guan2021real} also predict experimental observables (molecular toxicitiy or $^{13}$C NMR chemical shifts), but use 3D models that encode a single conformer that is optimized with xTB or MMFF94, respectively. Finally, \citet{van20243dreact} predict reaction activation barriers that were originally simulated at a high level of theory (CCSD(T)-F12a) using 3D ML models that encode (single) conformers of the reactant and product molecules, which were optimized with either xTB or DFT. Notably, whereas \citet{guan2021real} found that using MMFF94-level conformers instead of DFT-level conformers did not practically affect model accuracy in $^{13}$C chemical shift prediction, \citet{van20243dreact} found that in two tasks related to activation energy prediction, encoding DFT-level geometries improved model accuracy versus encoding xTB-level geometries.

\section{Methods}
\label{methods}

\subsection{Dataset overview.}

We use the dataset introduced by \citet{haas2024rapid} to obtain four ensemble-level regression targets. This dataset contains $>$8000 conformer ensembles of carboxylic acids optimized at the M06-2X/def2-TZVP // B3LYP-D3(BJ)/6-31G(d,p) level of theory. In summary, conformer ensembles of each acid were initially generated using MacroModel \citep{mohamadi1990macromodel} with a relative energy threshold of 5 kcal/mol before undergoing DFT-based geometry optimization. In the original study, ensembles with more than 20 conformers were automatically clustered, and no conformers were removed after geometry optimization regardless of their DFT-level energies post-optimization. In this study, we removed all molecules whose conformer ensembles underwent clustering or which contained conformers whose DFT-computed free energies post-optimization exceeded 5 kcal/mol relative to the lowest-energy conformation. This additional filtering was performed to reduce the likelihood that high-energy conformers or potentially missing conformations could add substantial noise to the computed ensemble-level properties. Our filtered dataset contains 5056 acids with their DFT-level conformer ensembles of at most 20 structures, labeled with descriptors for each conformer. Here, we only consider the Sterimol-L (max), Sterimol-L (min), Sterimol-B5 (max), and Sterimol-B5 (min) ensemble-level descriptors as machine learning regression targets. 

We also create two ``corrupted'' versions of the original DFT-level ensembles by taking each DFT-level conformer and separately re-optimizing its geometry with MMFF94 \citep{halgren1996merck} and GFN2-xTB \citep{bannwarth2019gfn2}. These conformer ensembles are essentially corrupted versions of the ground truth conformers, but are distinct from what would be produced using these inexpensive methods from scratch. 
Therefore, we additionally create \textit{inexpensive} conformer ensembles for each molecule by separately generating two new conformer ensembles optimized with either MMFF94 or GFN2-xTB. In both cases, we initially  use RDKit \citep{landrum2013rdkit} to enumerate up to 100 conformers with ETKDG \citep{riniker2015better}. Each conformer is then optimized with MMFF94. For the xTB-level ensembles, the conformers are further optimized with GFN2-xTB in the gas phase, and any xTB-optimized conformer having an energy greater than 5 kcal/mol relative to the lowest energy conformation is removed. Both the MMFF94- and xTB-level ensembles are then iteratively clustered with Butina clustering until each clustered ensemble contains fewer than 20 conformations.

For training and evaluating our ML models, we partition our filtered datasets into train, validation, and test sets containing 4056/500/1000 molecules, respectively. We randomly re-split the dataset three times and train models on each split to help ensure our analyses of model error are significant.

\subsection{Surrogate machine learning models.}

In this work, we make extensive use of 3D graph neural networks to predict conformer ensemble-level properties from encodings of MMFF94, xTB, and DFT-optimized molecular geometries. For consistency with \citet{haas2024rapid}, our experiments employ the DimeNet++ \citep{gasteiger2020fast} architecture, an E(3)-invariant 3D graph neural network that uses 2-hop pairwise atomic distances and 3-hop angles between triplets of atoms to learn molecular representations sensitive to 3D molecular geometry. 
Further descriptions of the DimeNet++ architecture can be found in its original paper. Here, we describe the encoding strategies that we use to learn representations of conformer ensembles.

\textbf{Encoding one conformation.}

We follow \citet{haas2024rapid}'s modeling strategy to predict an ensemble-level descriptor (Sterimol-L-min/max or Sterimol-B5-min/max) of the bond between the carbonyl- and $\alpha$-carbons of carboxlyic acids. The input to the model is a conformation described as a point cloud $\boldsymbol{P} = \{(\boldsymbol{r}_i, \boldsymbol{x}_i)\}_{i=1}^n$, where $\boldsymbol{r}_i \in \mathbb{R}^3$ are coordinates of atom $i$ and $\boldsymbol{x}_i \in \mathbb{R}^{d_x}$ are simple atomic features\footnote{Atom features include one-hot encodings of the element type, number of radical electrons, tetrahedral chirality, formal charge, ring size, degree, and total number of bonded hydrogen atoms. Except for the hydrogen on the -COOH functional group, hydrogens are treated implicitly.}. DimeNet++ is used to initially embed $\boldsymbol{P}$ into a set of learned atom representations $\{\boldsymbol{h}_i\}_{i=1}^n = \phi(\boldsymbol{P})$. We then predict the descriptor given the representations of the bonded atoms $a$ and $b$:
\begin{equation}
    \boldsymbol{h}_{\text{perm}} = f_{\text{perm}}(\boldsymbol{h}_a, \boldsymbol{h}_b) + f_{\text{perm}}(\boldsymbol{h}_b, \boldsymbol{h}_a)
\end{equation}
\begin{equation}
    \boldsymbol{h}_{\text{bond}} = (\boldsymbol{h}_{\text{perm}}, \sum_i \boldsymbol{h}_i)
    \label{eq:encoding}
\end{equation}
\begin{equation}
    \hat{y} = f_{\text{bond}}(\boldsymbol{h}_{\text{bond}})
\end{equation}
where $f_{\text{perm}}$ and $f_{\text{bond}}$ are multi-layer perceptrons (MLPs), $(\cdot \ ,\cdot)$ denotes concatenation in the feature dimension, and $\hat{y}\in \mathbb{R}$ is the predicted descriptor describing the bond between atoms $a$ and $b$.

\textbf{Encoding a set of conformations.}

We also consider models that encode a \textit{set} of conformers $\{\boldsymbol{P}_c\}_{c=1}^{n_c}$. These set encoders first embed each conformer individually into $\boldsymbol{h}^{(c)}_{\text{bond}}$ following Eq. \ref{eq:encoding}. They then encode an ensemble-level representation $\boldsymbol{h}_{\text{ensemble}}$ by linearly combining each $\boldsymbol{h}^{(c)}_{\text{bond}}$ with learned coefficients:
\begin{equation}
    \boldsymbol{h}_{\text{ensemble}} = \sum_{c=1}^{n_c} \alpha_c\ \boldsymbol{h}^{(c)}_{\text{bond}}
\end{equation}
\begin{equation}
    \tilde{\alpha}_c = f_{\text{gate}}(\boldsymbol{h}^{(c)}_{\text{bond}}) \ ; \ \ \ \ \ \ \ \alpha_c = \frac{\exp(\tilde{\alpha}_c)}{\sum_{c=1}^{n_c} \exp{(\tilde{\alpha}_c})}
\end{equation}
where $f_{\text{gate}}$ is an MLP mapping each $\boldsymbol{h}^{(c)}_{\text{bond}}$ to $\alpha_c \in \mathbb{R}$. Note that similar set encoding strategies have been previously explored \citep{zhu2023learning, axelrod2023molecular, chuang2020attention, zankov2023multi}. Using $\boldsymbol{h}_{\text{ensemble}}$, we then predict the target descriptor $\hat{y} = f_{\text{bond}}(\boldsymbol{h}_{\text{ensemble}})$.

\textbf{Overview of all trained models.}

In total, we train 14 types of 3D surrogate models to evaluate the impact of conformer quality on 3D representation learning of conformer ensembles (App. \ref{app:tabulatedmodels}). Each model is trained to predict the minimum or maximum Sterimol-L or B5 parameters of the ``ground-truth'' DFT-optimized conformer ensembles computed by \citet{haas2024rapid}; the models differ regarding the conformers they encode.

The first three types of models directly encode the ``active'' conformer whose Sterimol parameter is the minimum or maximum amongst the conformers in the DFT ensemble. The three models encode the active conformer in \textbf{(1)} its original DFT-optimized geometry, or its geometry following re-optimization (i.e., corruption) with \textbf{(2)} GFN2-xTB or \textbf{(3)} MMFF94.

The next three types of models also encode single conformations, but encode one \textit{random} conformation that is sampled from \textbf{(4)} the ground-truth DFT ensemble or the newly generated \textbf{(5)} xTB- or \textbf{(6)} MMFF94-optimized ensembles. The random conformation is presampled and fixed during training.

We also consider another three models that similarly encode random \textbf{(7)} DFT, \textbf{(8)} xTB, or \textbf{(9)} MMFF94-level conformers, but we train these models with conformer-based data augmentation. Prior to training, up to $n_c{=}10$ conformers are sampled from the full ensembles (we considered $n_c= 5, 10, 20$, but found no benefit beyond $n_c{=}10$). During each training iteration, one of these $n_c$ conformers is provided as input to the model. Notably, \citet{zhu2023learning} found that this strategy can slightly improve the prediction of ensemble-level properties without increasing inference costs. 

The next two models encode \textit{sets} of $n_c{=}10$ conformers sampled from the new \textbf{(10)} xTB- or \textbf{(11)} MMFF94-level ensembles. These $n_c$ sampled conformers are presampled and fixed during training.

Finally, the last three models also encode sets of $n_c$ conformers from either the new \textbf{(12)} MMFF94- or \textbf{(13)} xTB-level ensembles, \textit{or} \textbf{(14)} the original DFT-level ensembles. Crucially, one of the conformers is \textit{guaranteed} to be the ``active'' conformer that has been re-optimized at the same level of theory. We call these sets ``decoy-sets'' as they contain the active conformer and $n_c {-} 1$ decoys.

\subsection{Training and inference details.}

We separately train each model to predict each regression target. We retrain each model three times with different train/validation/test splits and report the average test-error across the three test sets. Each model is trained to convergence, and the best checkpoint is chosen based on the mean absolute error (MAE) on the corresponding validation set. We evaluate each model on the test set using the same encoding strategies that were used for training. For the models employing data augmentation, we average the models' predictions across all $n_c$ presampled conformations. 

\section{Results and Discussion}
\label{results_discussion}

\begin{figure}[t]
  \begin{center}
    \includegraphics[width=0.85\textwidth]{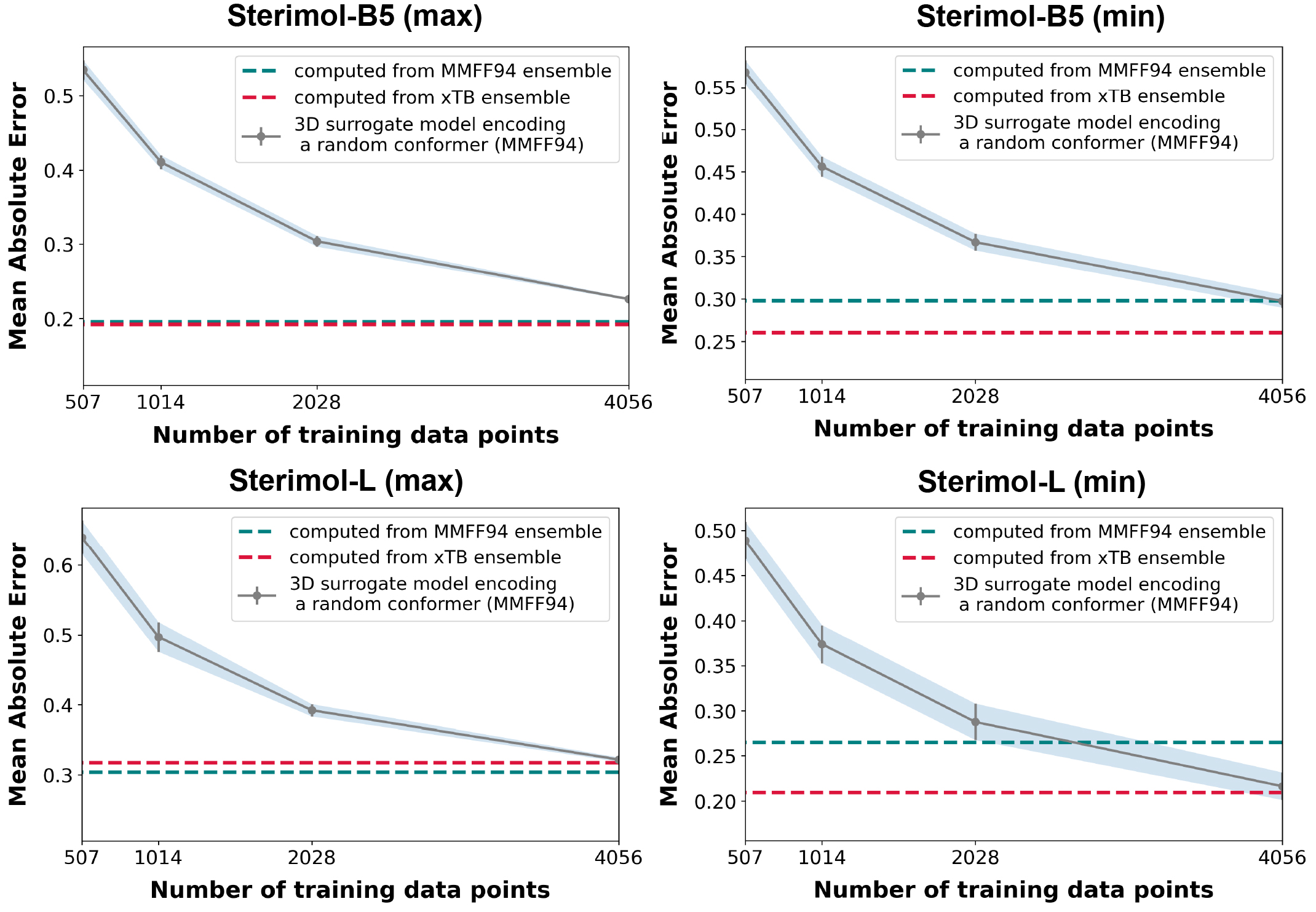}
  \end{center}
  \caption{Performance of 3D machine learning surrogate models trained to predict Sterimol-B5 (min/max) or Sterimol-L (min/max) descriptors of DFT-optimized conformer ensembles from encodings of random MMFF94-optimized conformers, as a function of training set size. Performance is measured by the mean absolute error on the test set, averaged across the three test sets. We compare model accuracy against the accuracy of simply computing these descriptors from cheap-to-simulate conformer ensembles optimized with either MMFF94 or xTB.}
  \label{fig:computed_descriptors_vs_surrogate}
\vspace{-10pt}
\end{figure}

\textbf{What
are the trade-offs between using ML surrogate models to predict properties of high-quality
ensembles versus just computing the properties from cheap conformer ensembles?}

We begin our analysis of the impact of conformer quality on machine learning surrogate modeling of conformer ensembles by first debating whether training such surrogate models is even necessary or cost-effective. Instead of training surrogate models to predict properties of high-quality ensembles, it may be easier to simply \textit{compute} these properties directly from cheap conformer ensembles that are optimized at lower-levels of theory. In our case, rather than \textit{predicting} the min/max Sterimol parameters of the DFT-optimized ensembles with a neural network, can we just compute these descriptors from MMFF94- or xTB-level ensembles? This strategy would avoid the costs of simulating \textit{any} DFT-level ensemble, as there would be no need to develop a dataset for neural network training.

Figure \ref{fig:computed_descriptors_vs_surrogate} plots the accuracy of surrogate ML models trained to predict min/max Sterimol-L and -B5 parameters of DFT-level ensembles from 3D encodings of random MMFF94 conformers. We contrast against the simple baseline of computing these descriptors from cheaply generated MMFF94- and xTB-optimized ensembles in order to highlight the potential disadvantages of indiscriminately training machine learning surrogate models to predict ensemble-level properties. Strikingly, even when training on over 4000 data points -- each corresponding to a DFT-optimized conformer ensemble that had to be simulated prior to model training -- these surrogate models are actually \textit{less} accurate than simply computing the descriptors from cheap ensembles, which requires no data collection at all. For these ensemble-level properties, therefore, training ML surrogate models is not worthwhile given (1) the high upfront costs of collecting DFT-quality training data, and (2) the little (if any) improvement in accuracy compared to just computing properties directly from cheap conformers.

Undoubtedly, these findings have multiple limitations that should be considered before drawing general conclusions regarding the value of training surrogate models to predict ensemble-level properties. First, the sensitivity of the property of interest to geometric quality is a key variable: Sterimol parameters describe the steric bulk around atoms or bonds, and are likely to be less sensitive to local geometric perturbations than (for instance) certain electronic properties, which may not be reliably computable from lower-quality geometries. Second, although simulating DFT-quality ensembles in order to train just \textit{one} surrogate model may not be cost-effective, if numerous descriptors need to be estimated, then the cost of training set calculation is effectively amortized. Relatedly, if the trained surrogate models will be used for ultra-large virtual screens, then training costs may be dwarfed by the cumulative cost of generating even relatively cheap conformer ensembles for numerous molecules. On the other hand, there are also intermediate options between training surrogate models on expensive DFT-level ensembles versus computing properties from cheap ensembles. For electronic properties sensitive to optimization strategy, for instance, can sufficient accuracies be obtained by using moderate geometry optimizations? If so, could surrogate models be trained on properties derived from these medium-level ensembles, for which large dataset generation is more accessible? Lastly, one should also consider the underlying imprecision in the high-level conformer ensembles themselves, especially imprecision originating from any undersampling or overclustering of the ensembles due to practical budget constraints. It may actually be \textit{more} accurate to compute properties from lower-level ensembles that can be more exhaustively enumerated, rather than from higher-level ensembles which may incidentally exclude important conformations.

\textbf{When encoding \textit{random} conformers to predict properties of higher quality conformer ensembles, does the geometric quality of the encoded conformation matter?}

Having established that alternative strategies beyond ML surrogate modeling may be more effective for estimating ensemble-level properties, we now pivot to discussing the impact of conformer quality on these models' learned representations of conformer ensembles. We start by first considering the types of models explored by \citet{haas2024rapid}, which encoded \textit{random} conformers optimized with MMFF94, a classical force field. When modeling properties of high-quality conformer ensembles with machine-learned representations of \textit{random} conformers, does conformer quality matter?

\begin{figure}[t]
  \begin{center}
    \includegraphics[width=0.8\textwidth]{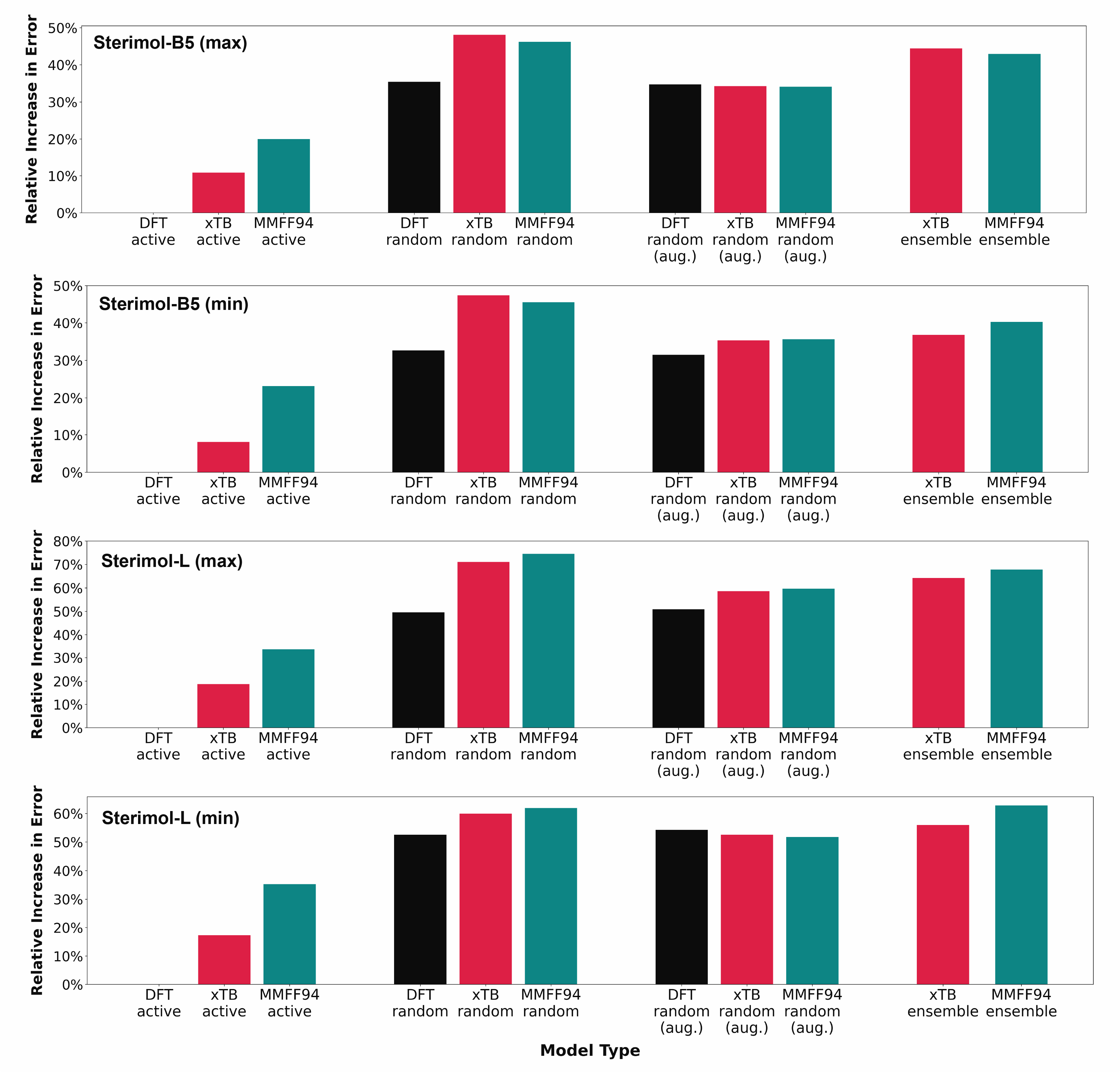}
  \end{center}
  \caption{Comparison of average prediction error across the test sets for different types of representation learning models that encode the true ``active'' conformation re-optimized with either DFT (black), xTB (red), or MMFF94 (teal); a single random conformer at various optimization levels; a single random conformer but training with data augmentation; or a set of random conformers at various optimization levels. Model performance is evaluated by the relative increase in error compared to the model that encodes the true DFT-level ``active'' conformation, which serves as an upper bound.}
  \label{fig:fidelity_comparisons}
\vspace{-10pt}
\end{figure}

We study this question by training 3D surrogate models that either encode a random MMFF94-, xTB-, or DFT-level conformer. To compare models, we compute the mean absolute error on our test set \textit{relative} to the error for a separate model that encodes the \textit{true} DFT-level ``active'' conformer, which gives an upper bound on performance given the quality of the training set. Figure \ref{fig:fidelity_comparisons} reports these errors, averaged across the three test sets. For all tasks, we find practically no difference in encoding MMFF94-level conformers instead of xTB-level conformers. This is somewhat surprising, as models which encode the \textit{active} conformer do show lower performance when that active conformer is corrupted with MMFF94 compared to xTB. When encoding \textit{random} conformers which are globally unrelated to the active conformer, therefore, local geometric quality seems to not matter.

Interestingly, we do find that encoding random DFT-level conformers does lead to slightly better performance compared to encoding xTB- or MMFF94-level conformers. However, this marginal benefit is negated when the models are trained with conformer-based data augmentation. We note that this observation agrees with what \citet{zhu2023learning} observed in other ensemble-level regression tasks: training with conformer data augmentation empirically makes 3D representation learning models more robust to local inaccuracies in conformational geometries.

\begin{figure}[t]
  \begin{center}
    \includegraphics[width=0.85\textwidth]{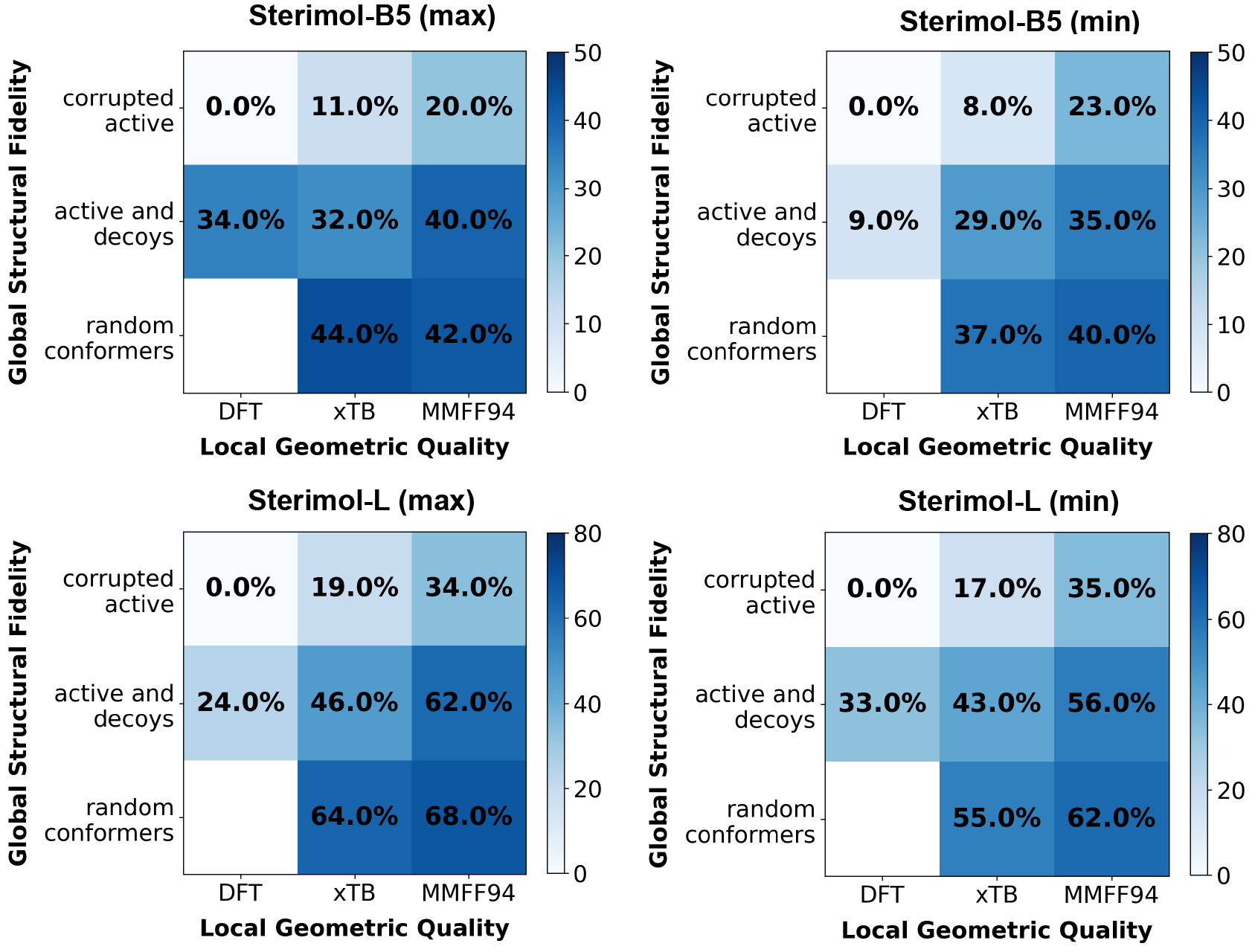}
  \end{center}
  \caption{Comparison in performance between ML surrogate models that encode the true ``active'' conformation that has been re-optimized (i.e., corrupted) with DFT, xTB, or MMFF94; models that encode ``decoy-sets'' containing the active conformer and up to 9 other decoys at the same level of theory; and models that encode sets of up to 10 random xTB- or MMFF94-optimized conformations. Performance is evaluated by the relative increase in MAE on the test set compared to the model that encodes the true DFT-level active conformation, and is averaged across three test sets.}
  \label{fig:ensemble_comparisons}
\vspace{-10pt}
\end{figure}

\textbf{How does the local geometric quality and global structural fidelity of an encoded \textit{set}
of conformers impact performance when predicting properties of
high-quality conformer ensembles?}

We now turn to our final inquiry: how does the local geometric quality and global structural fidelity of an encoded \textit{set} of conformers impact representation learning performance?
We begin by confirming others' observations regarding models that encode multiple conformers at once: explicitly encoding sets of conformers rarely leads to substantial (if any) performance improvements compared to encoding a single random conformation. Namely, in each of our property prediction regression tasks, encoding MMFF94- or xTB-optimized conformer ensembles performs \textit{worse} than encoding single random conformers while training with data augmentation (Fig. \ref{fig:fidelity_comparisons}). This finding is intuitively puzzling, as we might expect a set of conformers to be more information-rich compared to a single random conformer when predicting properties of conformer ensembles. Similar to before, we also find that there is not much benefit to optimizing these random conformers at relatively higher levels of theory (e.g., xTB vs. MMFF94). Empirically, then, if the encoded conformers are random conformers, conformer quality makes little difference, and it may be more prudent to just encode a single random conformer that is generated at a low level of theory -- at least for our particular regression tasks.

We now consider one hypothesis for why models encoding sets of conformers have yet to consistently  improve their simpler counterparts, especially when predicting properties associated with an ``active'' conformer: If the encoded set of conformers do not \textit{consistently} contain a \textit{close approximation} of the active conformation, then simultaneously encoding multiple conformations is unhelpful for property prediction. We explore this hypothesis by comparing the performance of our models which encode sets of \textit{random} MMFF94- or xTB-level conformers against models which encode artificially-constructed sets that are \textit{guaranteed} to contain an approximation of the true active conformer. Specifically, we construct models that encode so-called ``decoy-sets'', which include the (re-optimized) active conformer and up to 9 ``decoys'' that are sampled from either the DFT-, xTB-, or MMFF94-level ensembles.\footnote{In constructing these decoy-sets, it is important that the active and decoy conformers are optimized at the same level of theory so that the model cannot ``cheat'' by detecting the active conformer solely based on its local geometric quality, as opposed to its relevance to the target property.}\textsuperscript{,}\footnote{Although these decoy-sets are constructed artificially for our particular task of Sterimol parameter prediction, we stress the broad relevance of this scenario in other areas of chemical property prediction. For instance, it is commonly observed in protein-ligand docking that although the top-scoring docked pose may not be the true binding pose, the true pose is often contained within the top-k docking poses. Hence, obtaining sets of conformers that consistently contain the true ``active'' conformation is not unrealistic.}
In constructing these decoy-sets, we seek to analyze three phenomena: 
\begin{enumerate}
    \item the change in performance caused by encoding decoys alongside the (uncorrupted) active;
    \item the change in performance caused by corrupting the decoy-sets via low-level re-optimization;
    \item the difference in performance between models encoding corrupted decoy-sets versus those encoding \textit{random} conformers that aren't guaranteed to include the active conformer.
\end{enumerate}
Overall, these decoy-sets are intended to simulate the incremental loss of perfect structural information (related to both local geometric quality \textit{and} global structural fidelity) that is incurred by encoding sets of random low-level conformers instead of the ground-truth conformer ensemble itself.

Across all tasks, we first  observe that relative to models that encode the true DFT-optimized active conformer, simultaneously encoding the DFT-level active along with DFT-level decoys substantially reduces model performance by $\sim$10-35\% (Fig. \ref{fig:ensemble_comparisons}). 
One one hand, we might expect that additionally encoding decoys would reduce \textit{some} performance compared to just encoding the true active. We note that this particular model still has access to perfect information to solve the task, as the active conformer could in principle be identified among the encoded conformers by its minimum or maximum Sterimol parameter. Hence, just diluting the structural information provided to the model substantially reduces model performance, even without decreasing the local geometric quality \textit{or} global structural fidelity of the encoded conformers. 
Preventing this performance loss by designing neural architectures to better distinguish ``active'' vs. ``decoy'' conformers could serve as a concrete goal of future efforts seeking to improve the performance of models encoding sets of conformers. 

Finally, we analyze the impact of decreasing the local geometric quality and global structural fidelity of the encoded decoy-sets (Fig. \ref{fig:ensemble_comparisons}). Crucially, as the active and decoy conformers are corrupted via re-optimization with decreasing levels of theory (e.g., xTB and MMFF94), model performance progressively degrades. Performance also always degrades when encoding sets of \textit{random} conformers compared to sets that explicitly include the active conformer. Interestingly, nearly all performance benefits associated with including the active conformer in the encoded decoy-sets are lost when the active is sufficiently corrupted; models encoding MMFF94-level decoy-sets perform similarly compared to models encoding \textit{random} xTB-level conformers, and only marginally better than models encoding random MMFF94-level conformers. Overall, this suggests that when modeling ensemble-level properties dependent on (unknown) active conformers, encoding multiple conformers at once is only worthwhile (compared to encoding random conformers) if an encoded conformer closely approximates the true active, both in terms of its global structure and its local geometric quality.




\section{Summary}
\label{conclusion}

Training machine learning surrogate models to predict properties of conformer ensembles offers a promising strategy to expand conformational analyses to large chemical spaces that would be impractical to directly simulate at high levels of theory. However, choosing the type of surrogate model to maximize predictive performance opens a host of questions regarding the ability of 3D representation learning models to learn rich representations of high-quality conformer ensembles when only lower-quality molecular structures are readily accessible. When encoding random conformations as the input molecular representation, does conformer quality matter? When using models that encode sets of conformers to predict properties dependent on a certain ``active'' conformation, how does the presence and geometric quality of the active conformer within the encoded ensemble affect model performance? And perhaps controversially, is it even worthwhile to train machine learning surrogate models for this purpose, or is it more effective to just compute the properties of interest directly from lower-quality conformer ensembles that are far cheaper to simulate? 

For our ensemble-level properties that depend on a certain active conformation, we demonstrate:
\begin{itemize}
    \item It can actually be more cost-effective \textit{and} accurate to simply compute the properties from cheap-to-simulate conformer ensembles rather than training ML surrogate models, which requires computing high-quality conformer ensembles for numerous training data points.
    \item When training ML surrogate models that encode \textit{random} conformers to predict properties of high-quality ensembles, the optimization level of the encoded conformers does not matter.
    \item Encoding \textit{sets} of random conformers does not improve upon simpler models that only encode single conformers. The relatively poor performance of such models can be (in part) attributed to (1) non-robustness to ``decoy'' conformations that are encoded alongside an approximation of the true active; (2) the relatively poor local geometric quality of the encoded conformers relative to the true active; and (3) the poor global structural fidelity of the encoded conformers, i.e., whether an encoded conformer matches the active conformation.
    \end{itemize}
Although our analyses and results are specific to our particular case study on Sterimol parameter prediction, we expect such considerations to be relevant to similar problem settings in chemical property prediction that are historically approached via expensive conformational analysis. 

\newpage

\begin{ack}
The authors thank Shree Sowndarya, Brittany Haas, Melissa Hardy, Robert Paton, and Matthew Sigman for helpful discussions. This material is based upon work supported by the National Science Foundation Graduate Research Fellowship under Grant No. 2141064 and the by the National Science Foundation under the CCI Center for Computer-Assisted Synthesis (CHE-2202693). The authors acknowledge the MIT SuperCloud and Lincoln Laboratory Supercomputing Center for providing HPC resources that have contributed to the research results reported within this paper.
\end{ack}

\section*{Reproducibility Statement}
All code and data needed to reproduce the results reported within this paper are available at \url{https://github.com/keiradams/conformer_ensemble_quality}.

\bibliography{main}

\begin{thebibliography}{53}
\providecommand{\natexlab}[1]{#1}
\providecommand{\url}[1]{\texttt{#1}}
\expandafter\ifx\csname urlstyle\endcsname\relax
  \providecommand{\doi}[1]{doi: #1}\else
  \providecommand{\doi}{doi: \begingroup \urlstyle{rm}\Url}\fi

\bibitem[Anstine et~al.(2024)Anstine, Zubatyuk, and Isayev]{anstine2024aimnet2}
Dylan Anstine, Roman Zubatyuk, and Olexandr Isayev.
\newblock Aimnet2: a neural network potential to meet your neutral, charged, organic, and elemental-organic needs.
\newblock 2024.

\bibitem[Axelrod \& Gomez-Bombarelli(2023)Axelrod and Gomez-Bombarelli]{axelrod2023molecular}
Simon Axelrod and Rafael Gomez-Bombarelli.
\newblock Molecular machine learning with conformer ensembles.
\newblock \emph{Machine Learning: Science and Technology}, 4\penalty0 (3):\penalty0 035025, 2023.

\bibitem[Bannwarth et~al.(2019)Bannwarth, Ehlert, and Grimme]{bannwarth2019gfn2}
Christoph Bannwarth, Sebastian Ehlert, and Stefan Grimme.
\newblock Gfn2-xtb—an accurate and broadly parametrized self-consistent tight-binding quantum chemical method with multipole electrostatics and density-dependent dispersion contributions.
\newblock \emph{Journal of chemical theory and computation}, 15\penalty0 (3):\penalty0 1652--1671, 2019.

\bibitem[Besora et~al.(2011)Besora, Braga, Ujaque, Maseras, and Lled{\'o}s]{besora2011importance}
Maria Besora, Ataualpa~AC Braga, Gregori Ujaque, Feliu Maseras, and Agust{\'\i} Lled{\'o}s.
\newblock The importance of conformational search: a test case on the catalytic cycle of the suzuki--miyaura cross-coupling.
\newblock \emph{Theoretical Chemistry Accounts}, 128:\penalty0 639--646, 2011.

\bibitem[Brethome et~al.(2019)Brethome, Fletcher, and Paton]{brethome2019conformational}
Alexandre~V Brethome, Stephen~P Fletcher, and Robert~S Paton.
\newblock Conformational effects on physical-organic descriptors: the case of sterimol steric parameters.
\newblock \emph{ACS catalysis}, 9\penalty0 (3):\penalty0 2313--2323, 2019.

\bibitem[Chuang \& Keiser(2020)Chuang and Keiser]{chuang2020attention}
Kangway~V Chuang and Michael~J Keiser.
\newblock Attention-based learning on molecular ensembles.
\newblock \emph{arXiv preprint arXiv:2011.12820}, 2020.

\bibitem[Cremer et~al.(2023)Cremer, Medrano~Sandonas, Tkatchenko, Clevert, and De~Fabritiis]{cremer2023equivariant}
Julian Cremer, Leonardo Medrano~Sandonas, Alexandre Tkatchenko, Djork-Arn{\'e} Clevert, and Gianni De~Fabritiis.
\newblock Equivariant graph neural networks for toxicity prediction.
\newblock \emph{Chemical Research in Toxicology}, 36\penalty0 (10):\penalty0 1561--1573, 2023.

\bibitem[Davies et~al.(2012)Davies, Planas, and Rovira]{davies2012conformational}
Gideon~J Davies, Antoni Planas, and Carme Rovira.
\newblock Conformational analyses of the reaction coordinate of glycosidases.
\newblock \emph{Accounts of chemical research}, 45\penalty0 (2):\penalty0 308--316, 2012.

\bibitem[Dietterich et~al.(1997)Dietterich, Lathrop, and Lozano-P{\'e}rez]{dietterich1997solving}
Thomas~G Dietterich, Richard~H Lathrop, and Tom{\'a}s Lozano-P{\'e}rez.
\newblock Solving the multiple instance problem with axis-parallel rectangles.
\newblock \emph{Artificial intelligence}, 89\penalty0 (1-2):\penalty0 31--71, 1997.

\bibitem[Forli et~al.(2016)Forli, Huey, Pique, Sanner, Goodsell, and Olson]{forli2016computational}
Stefano Forli, Ruth Huey, Michael~E Pique, Michel~F Sanner, David~S Goodsell, and Arthur~J Olson.
\newblock Computational protein--ligand docking and virtual drug screening with the autodock suite.
\newblock \emph{Nature protocols}, 11\penalty0 (5):\penalty0 905--919, 2016.

\bibitem[Friedrich et~al.(2017)Friedrich, de~Bruyn~Kops, Flachsenberg, Sommer, Rarey, and Kirchmair]{friedrich2017benchmarking}
Nils-Ole Friedrich, Christina de~Bruyn~Kops, Florian Flachsenberg, Kai Sommer, Matthias Rarey, and Johannes Kirchmair.
\newblock Benchmarking commercial conformer ensemble generators.
\newblock \emph{Journal of chemical information and modeling}, 57\penalty0 (11):\penalty0 2719--2728, 2017.

\bibitem[Friesner et~al.(2004)Friesner, Banks, Murphy, Halgren, Klicic, Mainz, Repasky, Knoll, Shelley, Perry, et~al.]{friesner2004glide}
Richard~A Friesner, Jay~L Banks, Robert~B Murphy, Thomas~A Halgren, Jasna~J Klicic, Daniel~T Mainz, Matthew~P Repasky, Eric~H Knoll, Mee Shelley, Jason~K Perry, et~al.
\newblock Glide: a new approach for rapid, accurate docking and scoring. 1. method and assessment of docking accuracy.
\newblock \emph{Journal of medicinal chemistry}, 47\penalty0 (7):\penalty0 1739--1749, 2004.

\bibitem[Ganea et~al.(2021)Ganea, Pattanaik, Coley, Barzilay, Jensen, Green, and Jaakkola]{ganea2021geomol}
Octavian Ganea, Lagnajit Pattanaik, Connor Coley, Regina Barzilay, Klavs Jensen, William Green, and Tommi Jaakkola.
\newblock Geomol: Torsional geometric generation of molecular 3d conformer ensembles.
\newblock \emph{Advances in Neural Information Processing Systems}, 34:\penalty0 13757--13769, 2021.

\bibitem[Gasteiger et~al.(2020)Gasteiger, Giri, Margraf, and G{\"u}nnemann]{gasteiger2020fast}
Johannes Gasteiger, Shankari Giri, Johannes~T Margraf, and Stephan G{\"u}nnemann.
\newblock Fast and uncertainty-aware directional message passing for non-equilibrium molecules.
\newblock \emph{arXiv preprint arXiv:2011.14115}, 2020.

\bibitem[Gensch et~al.(2022)Gensch, dos Passos~Gomes, Friederich, Peters, Gaudin, Pollice, Jorner, Nigam, Lindner-D’Addario, Sigman, et~al.]{gensch2022comprehensive}
Tobias Gensch, Gabriel dos Passos~Gomes, Pascal Friederich, Ellyn Peters, Th{\'e}ophile Gaudin, Robert Pollice, Kjell Jorner, AkshatKumar Nigam, Michael Lindner-D’Addario, Matthew~S Sigman, et~al.
\newblock A comprehensive discovery platform for organophosphorus ligands for catalysis.
\newblock \emph{Journal of the American Chemical Society}, 144\penalty0 (3):\penalty0 1205--1217, 2022.

\bibitem[Gilson \& Zhou(2007)Gilson and Zhou]{gilson2007calculation}
Michael~K Gilson and Huan-Xiang Zhou.
\newblock Calculation of protein-ligand binding affinities.
\newblock \emph{Annu. Rev. Biophys. Biomol. Struct.}, 36\penalty0 (1):\penalty0 21--42, 2007.

\bibitem[Guan et~al.(2018)Guan, Ingman, Rooks, and Wheeler]{guan2018aaron}
Yanfei Guan, Victoria~M Ingman, Benjamin~J Rooks, and Steven~E Wheeler.
\newblock Aaron: an automated reaction optimizer for new catalysts.
\newblock \emph{Journal of chemical theory and computation}, 14\penalty0 (10):\penalty0 5249--5261, 2018.

\bibitem[Guan et~al.(2021)Guan, Sowndarya, Gallegos, John, and Paton]{guan2021real}
Yanfei Guan, SV~Shree Sowndarya, Liliana~C Gallegos, Peter C~St John, and Robert~S Paton.
\newblock Real-time prediction of 1 h and 13 c chemical shifts with dft accuracy using a 3d graph neural network.
\newblock \emph{Chemical Science}, 12\penalty0 (36):\penalty0 12012--12026, 2021.

\bibitem[Haas et~al.(2024)Haas, Hardy, SV, Adams, Coley, Paton, and Sigman]{haas2024rapid}
Brittany Haas, Melissa Hardy, Shree~Sowndarya SV, Keir Adams, Connor Coley, Robert Paton, and Matthew Sigman.
\newblock Rapid prediction of conformationally-dependent dft-level descriptors using graph neural networks for carboxylic acids and alkyl amines.
\newblock 2024.

\bibitem[Haas et~al.(2022)Haas, Goetz, Bahamonde, McWilliams, and Sigman]{haas2022predicting}
Brittany~C Haas, Adam~E Goetz, Ana Bahamonde, J~Christopher McWilliams, and Matthew~S Sigman.
\newblock Predicting relative efficiency of amide bond formation using multivariate linear regression.
\newblock \emph{Proceedings of the National Academy of Sciences}, 119\penalty0 (16):\penalty0 e2118451119, 2022.

\bibitem[Halgren(1996)]{halgren1996merck}
Thomas~A Halgren.
\newblock Merck molecular force field. i. basis, form, scope, parameterization, and performance of mmff94.
\newblock \emph{Journal of computational chemistry}, 17\penalty0 (5-6):\penalty0 490--519, 1996.

\bibitem[Hansen et~al.(2016)Hansen, Rosales, Tutkowski, Norrby, and Wiest]{hansen2016prediction}
Eric Hansen, Anthony~R Rosales, Brandon Tutkowski, Per-Ola Norrby, and Olaf Wiest.
\newblock Prediction of stereochemistry using q2mm.
\newblock \emph{Accounts of Chemical Research}, 49\penalty0 (5):\penalty0 996--1005, 2016.

\bibitem[Hawkins(2017)]{hawkins2017conformation}
Paul~CD Hawkins.
\newblock Conformation generation: the state of the art.
\newblock \emph{Journal of chemical information and modeling}, 57\penalty0 (8):\penalty0 1747--1756, 2017.

\bibitem[Hawkins et~al.(2010)Hawkins, Skillman, Warren, Ellingson, and Stahl]{hawkins2010conformer}
Paul~CD Hawkins, A~Geoffrey Skillman, Gregory~L Warren, Benjamin~A Ellingson, and Matthew~T Stahl.
\newblock Conformer generation with omega: algorithm and validation using high quality structures from the protein databank and cambridge structural database.
\newblock \emph{Journal of chemical information and modeling}, 50\penalty0 (4):\penalty0 572--584, 2010.

\bibitem[Howard \& Kollman(1988)Howard and Kollman]{howard1988analysis}
Allison~E Howard and Peter~A Kollman.
\newblock An analysis of current methodologies for conformational searching of complex molecules.
\newblock \emph{Journal of Medicinal Chemistry}, 31\penalty0 (9):\penalty0 1669--1675, 1988.

\bibitem[Ilse et~al.(2018)Ilse, Tomczak, and Welling]{ilse2018attention}
Maximilian Ilse, Jakub Tomczak, and Max Welling.
\newblock Attention-based deep multiple instance learning.
\newblock In \emph{International conference on machine learning}, pp.\  2127--2136. PMLR, 2018.

\bibitem[Jacobson et~al.(2017)Jacobson, Bochevarov, Watson, Hughes, Rinaldo, Ehrlich, Steinbrecher, Vaitheeswaran, Philipp, Halls, et~al.]{jacobson2017automated}
Leif~D Jacobson, Art~D Bochevarov, Mark~A Watson, Thomas~F Hughes, David Rinaldo, Stephan Ehrlich, Thomas~B Steinbrecher, S~Vaitheeswaran, Dean~M Philipp, Mathew~D Halls, et~al.
\newblock Automated transition state search and its application to diverse types of organic reactions.
\newblock \emph{Journal of chemical theory and computation}, 13\penalty0 (11):\penalty0 5780--5797, 2017.

\bibitem[Jing et~al.(2022)Jing, Corso, Chang, Barzilay, and Jaakkola]{jing2022torsional}
Bowen Jing, Gabriele Corso, Jeffrey Chang, Regina Barzilay, and Tommi Jaakkola.
\newblock Torsional diffusion for molecular conformer generation.
\newblock \emph{Advances in Neural Information Processing Systems}, 35:\penalty0 24240--24253, 2022.

\bibitem[Kim et~al.(2017)Kim, Jang, Yadav, and Kim]{kim2017comparison}
Hyoungrae Kim, Cheongyun Jang, Dharmendra~K Yadav, and Mi-hyun Kim.
\newblock The comparison of automated clustering algorithms for resampling representative conformer ensembles with rmsd matrix.
\newblock \emph{Journal of cheminformatics}, 9:\penalty0 1--26, 2017.

\bibitem[Kuntz et~al.(1982)Kuntz, Blaney, Oatley, Langridge, and Ferrin]{kuntz1982geometric}
Irwin~D Kuntz, Jeffrey~M Blaney, Stuart~J Oatley, Robert Langridge, and Thomas~E Ferrin.
\newblock A geometric approach to macromolecule-ligand interactions.
\newblock \emph{Journal of molecular biology}, 161\penalty0 (2):\penalty0 269--288, 1982.

\bibitem[Landrum(2013)]{landrum2013rdkit}
Greg Landrum.
\newblock Rdkit documentation.
\newblock \emph{Release}, 1\penalty0 (1-79):\penalty0 4, 2013.

\bibitem[Laplaza et~al.(2024)Laplaza, Wodrich, and Corminboeuf]{laplaza2024overcoming}
Ruben Laplaza, Matthew~D Wodrich, and Clemence Corminboeuf.
\newblock Overcoming the pitfalls of computing reaction selectivity from ensembles of transition states.
\newblock \emph{The Journal of Physical Chemistry Letters}, 15\penalty0 (29):\penalty0 7363--7370, 2024.

\bibitem[Liu et~al.(2021)Liu, Fu, Zhang, Wang, Xie, Yuan, Luo, Xu, Xu, and Ji]{liu2021fast}
Meng Liu, Cong Fu, Xuan Zhang, Limei Wang, Yaochen Xie, Hao Yuan, Youzhi Luo, Zhao Xu, Shenglong Xu, and Shuiwang Ji.
\newblock Fast quantum property prediction via deeper 2d and 3d graph networks.
\newblock \emph{arXiv preprint arXiv:2106.08551}, 2021.

\bibitem[Liu et~al.(2022)Liu, Zubatiuk, Roitberg, and Isayev]{liu2022auto3d}
Zhen Liu, Tetiana Zubatiuk, Adrian Roitberg, and Olexandr Isayev.
\newblock Auto3d: Automatic generation of the low-energy 3d structures with ani neural network potentials.
\newblock \emph{Journal of Chemical Information and Modeling}, 62\penalty0 (22):\penalty0 5373--5382, 2022.

\bibitem[Maron \& Lozano-P{\'e}rez(1997)Maron and Lozano-P{\'e}rez]{maron1997framework}
Oded Maron and Tom{\'a}s Lozano-P{\'e}rez.
\newblock A framework for multiple-instance learning.
\newblock \emph{Advances in neural information processing systems}, 10, 1997.

\bibitem[Meng et~al.(2011)Meng, Zhang, Mezei, and Cui]{meng2011molecular}
Xuan-Yu Meng, Hong-Xing Zhang, Mihaly Mezei, and Meng Cui.
\newblock Molecular docking: a powerful approach for structure-based drug discovery.
\newblock \emph{Current computer-aided drug design}, 7\penalty0 (2):\penalty0 146--157, 2011.

\bibitem[Mezei \& Beveridge(1986)Mezei and Beveridge]{mezei1986free}
M~Mezei and DL~Beveridge.
\newblock Free energy simulations.
\newblock \emph{Annals of the New York Academy of Sciences}, 482\penalty0 (1):\penalty0 1--23, 1986.

\bibitem[Mohamadi et~al.(1990)Mohamadi, Richards, Guida, Liskamp, Lipton, Caufield, Chang, Hendrickson, and Still]{mohamadi1990macromodel}
Fariborz Mohamadi, Nigel~GJ Richards, Wayne~C Guida, Rob Liskamp, Mark Lipton, Craig Caufield, George Chang, Thomas Hendrickson, and W~Clark Still.
\newblock Macromodel—an integrated software system for modeling organic and bioorganic molecules using molecular mechanics.
\newblock \emph{Journal of Computational Chemistry}, 11\penalty0 (4):\penalty0 440--467, 1990.

\bibitem[Perola \& Charifson(2004)Perola and Charifson]{perola2004conformational}
Emanuele Perola and Paul~S Charifson.
\newblock Conformational analysis of drug-like molecules bound to proteins: an extensive study of ligand reorganization upon binding.
\newblock \emph{Journal of medicinal chemistry}, 47\penalty0 (10):\penalty0 2499--2510, 2004.

\bibitem[Reiser et~al.(2022)Reiser, Neubert, Eberhard, Torresi, Zhou, Shao, Metni, van Hoesel, Schopmans, Sommer, et~al.]{reiser2022graph}
Patrick Reiser, Marlen Neubert, Andr{\'e} Eberhard, Luca Torresi, Chen Zhou, Chen Shao, Houssam Metni, Clint van Hoesel, Henrik Schopmans, Timo Sommer, et~al.
\newblock Graph neural networks for materials science and chemistry.
\newblock \emph{Communications Materials}, 3\penalty0 (1):\penalty0 93, 2022.

\bibitem[Riniker \& Landrum(2015)Riniker and Landrum]{riniker2015better}
Sereina Riniker and Gregory~A Landrum.
\newblock Better informed distance geometry: using what we know to improve conformation generation.
\newblock \emph{Journal of chemical information and modeling}, 55\penalty0 (12):\penalty0 2562--2574, 2015.

\bibitem[Rosales et~al.(2019)Rosales, Wahlers, Lim{\'e}, Meadows, Leslie, Savin, Bell, Hansen, Helquist, Munday, et~al.]{rosales2019rapid}
Anthony~R Rosales, Jessica Wahlers, Elaine Lim{\'e}, Rebecca~E Meadows, Kevin~W Leslie, Rhona Savin, Fiona Bell, Eric Hansen, Paul Helquist, Rachel~H Munday, et~al.
\newblock Rapid virtual screening of enantioselective catalysts using catvs.
\newblock \emph{Nature Catalysis}, 2\penalty0 (1):\penalty0 41--45, 2019.

\bibitem[Smith et~al.(2017)Smith, Isayev, and Roitberg]{smith2017ani}
Justin~S Smith, Olexandr Isayev, and Adrian~E Roitberg.
\newblock Ani-1: an extensible neural network potential with dft accuracy at force field computational cost.
\newblock \emph{Chemical science}, 8\penalty0 (4):\penalty0 3192--3203, 2017.

\bibitem[Unzueta et~al.(2021)Unzueta, Greenwell, and Beran]{unzueta2021predicting}
Pablo~A Unzueta, Chandler~S Greenwell, and Gregory~JO Beran.
\newblock Predicting density functional theory-quality nuclear magnetic resonance chemical shifts via $\delta$-machine learning.
\newblock \emph{Journal of Chemical Theory and Computation}, 17\penalty0 (2):\penalty0 826--840, 2021.

\bibitem[van Gerwen et~al.(2024)van Gerwen, Briling, Bunne, Somnath, Laplaza, Krause, and Corminboeuf]{van20243dreact}
Puck van Gerwen, Ksenia~R Briling, Charlotte Bunne, Vignesh~Ram Somnath, Ruben Laplaza, Andreas Krause, and Clemence Corminboeuf.
\newblock 3dreact: Geometric deep learning for chemical reactions.
\newblock \emph{Journal of Chemical Information and Modeling}, 64\penalty0 (15):\penalty0 5771--5785, 2024.

\bibitem[Weinreich et~al.(2021)Weinreich, Browning, and von Lilienfeld]{weinreich2021machine}
Jan Weinreich, Nicholas~J Browning, and O~Anatole von Lilienfeld.
\newblock Machine learning of free energies in chemical compound space using ensemble representations: Reaching experimental uncertainty for solvation.
\newblock \emph{The Journal of Chemical Physics}, 154\penalty0 (13), 2021.

\bibitem[Yongye et~al.(2010)Yongye, Bender, and Mart{\'\i}nez-Mayorga]{yongye2010dynamic}
Austin~B Yongye, Andreas Bender, and Karina Mart{\'\i}nez-Mayorga.
\newblock Dynamic clustering threshold reduces conformer ensemble size while maintaining a biologically relevant ensemble.
\newblock \emph{Journal of computer-aided molecular design}, 24:\penalty0 675--686, 2010.

\bibitem[Zahrt et~al.(2019)Zahrt, Henle, Rose, Wang, Darrow, and Denmark]{zahrt2019prediction}
Andrew~F Zahrt, Jeremy~J Henle, Brennan~T Rose, Yang Wang, William~T Darrow, and Scott~E Denmark.
\newblock Prediction of higher-selectivity catalysts by computer-driven workflow and machine learning.
\newblock \emph{Science}, 363\penalty0 (6424):\penalty0 eaau5631, 2019.

\bibitem[Zankov et~al.(2023)Zankov, Madzhidov, Polishchuk, Sidorov, and Varnek]{zankov2023multi}
Dmitry Zankov, Timur Madzhidov, Pavel Polishchuk, Pavel Sidorov, and Alexandre Varnek.
\newblock Multi-instance learning approach to the modeling of enantioselectivity of conformationally flexible organic catalysts.
\newblock \emph{Journal of Chemical Information and Modeling}, 63\penalty0 (21):\penalty0 6629--6641, 2023.

\bibitem[Zankov et~al.(2024)Zankov, Madzhidov, Varnek, and Polishchuk]{zankov2024chemical}
Dmitry Zankov, Timur Madzhidov, Alexandre Varnek, and Pavel Polishchuk.
\newblock Chemical complexity challenge: Is multi-instance machine learning a solution?
\newblock \emph{Wiley Interdisciplinary Reviews: Computational Molecular Science}, 14\penalty0 (1):\penalty0 e1698, 2024.

\bibitem[Zankov et~al.(2021)Zankov, Matveieva, Nikonenko, Nugmanov, Baskin, Varnek, Polishchuk, and Madzhidov]{zankov2021qsar}
Dmitry~V Zankov, Mariia Matveieva, Aleksandra~V Nikonenko, Ramil~I Nugmanov, Igor~I Baskin, Alexandre Varnek, Pavel Polishchuk, and Timur~I Madzhidov.
\newblock Qsar modeling based on conformation ensembles using a multi-instance learning approach.
\newblock \emph{Journal of chemical information and modeling}, 61\penalty0 (10):\penalty0 4913--4923, 2021.

\bibitem[Zhu et~al.(2023)Zhu, Hwang, Adams, Liu, Nan, Stenfors, Du, Chauhan, Wiest, Isayev, et~al.]{zhu2023learning}
Yanqiao Zhu, Jeehyun Hwang, Keir Adams, Zhen Liu, Bozhao Nan, Brock Stenfors, Yuanqi Du, Jatin Chauhan, Olaf Wiest, Olexandr Isayev, et~al.
\newblock Learning over molecular conformer ensembles: Datasets and benchmarks.
\newblock In \emph{The Twelfth International Conference on Learning Representations}, 2023.

\bibitem[Zubatyuk et~al.(2019)Zubatyuk, Smith, Leszczynski, and Isayev]{zubatyuk2019accurate}
Roman Zubatyuk, Justin~S Smith, Jerzy Leszczynski, and Olexandr Isayev.
\newblock Accurate and transferable multitask prediction of chemical properties with an atoms-in-molecules neural network.
\newblock \emph{Science advances}, 5\penalty0 (8):\penalty0 eaav6490, 2019.

\end{thebibliography}
\bibliographystyle{main}

\clearpage
\appendix

\section{Appendix}

\ifthenelse{\boolean{includeBackgroundOnHaas}}{

\subsection{Additional background on \citet{haas2024rapid}}
\label{app:haas}
\textit{Parts of the following sections have previously appeared in ... The excerpts included here are among my own direct contributions to \citet{haas2024rapid}.}

Computing large DFT-level descriptor libraries for small molecules is an established strategy in data-driven chemistry to facilitate downstream quantitative analyses of chemical reactivity. However, for ultra-large chemical spaces, computing steric, electronic, and/or stereoelectronic descriptors of DFT-optimized conformer ensembles can be computationally prohibitive due to the costs of traditional conformational analysis, necessitating an alternative approach to descriptor estimation. In \citet{haas2024rapid}, we trained machine learning property prediction models based on graph neural networks in order to predict over 80 ensemble-level descriptors of carboxylic acids, primary amines, and secondary amines, using over $16{,}000$ newly generated conformer ensembles computed at the DFT-level as training data. These ensemble-level descriptors include the minimum, maximum, and Boltzmann-averaged properties of the conformers in each ensemble, as well as the properties of the lowest-energy conformation in each ensemble. Properties include molecular-level properties such as frontier molecular orbital (HOMO/LUMO) energies and dipole moments; atom-level properties such as natural bond orbital (NBO) parameters and NMR chemical shieldings; and bond-level properties like Sterimol parameters and IR frequencies. Acids and amines were chosen as the target chemical systems due to their high relevane to medicinal chemistry, as evidenced by the prevalence of amide coupling reactions in the syntheses of small molecule drug candidates.

In general, the trained 2D graph neural networks -- despite having no explicit 3D awareness -- achieved high accuracy, with $R^2$ values often exceeding 0.9. Nevertheless, certain descriptors -- particularly those that are ultra-sensitive to molecular geometry and/or conformational flexibility such as Sterimol parameters and NBO partial charges -- were predicted with notably less accuracy. This motivated the exploration of machine learning models that have greater 3D expressivity than standard graph neural networks, which only encode a (potentially featurized) molecular graph.

The following sections describe how we improved the accuracy of these 2D surrogate models by training 3D graph neural networks to predict the descriptors of the DFT-level conformer ensembles from encodings of low-level (and therefore cheap-to-simulate) conformers optimized with MMFF94.

\subsubsection{Incorporating 3D geometries for descriptor prediction}
In contrast to 2D GNNs, which solely encode the information stored in a graph representation of a molecule and pass messages between covalently bonded atoms, 3D GNNs generally encode point clouds representing the coordinates of a specific 3D conformation of the molecule. Under this paradigm, edges in the 3D graph no longer solely correspond to covalent bonds, but additionally include non-covalent edges between any two atoms separated by a distance within a defined cutoff. By explicitly encoding the local geometry of these 3D graphs (e.g., in terms of distances, angles, and dihedrals), 3D GNNs have the capacity to directly capture 3D geometric features present in the conformers provided as input to the model. We hypothesized that this enhanced geometric expressivity would improve the surrogate descriptor prediction models, particularly for geometry-sensitive descriptors such as Sterimol values and NBO partial charges.

The requirement of input conformers when using 3D GNNs introduces additional questions with regards to how those conformers should be generated. Applying DFT-level conformers, such as those used to compute the target acid/amine descriptors, would offset the desired cost savings of using a ML surrogate, so we instead employed cheap-to-acquire conformers that are separately embedded with the ETKDG algorithm and optimized with MMFF94, all via RDKit. For the relatively low molecular weight acids and amines studied here, this conformer generation was fast and accessible for thousands of molecules with limited computational resources (ca. 2.5 CPU seconds per molecule). Our use of MMFF94 conformers assumes that these structures provide an adequate representation of the ground-truth DFT-optimized conformer ensemble, which we show to be empirically defensible, albeit imperfect. For instance, the cheap surrogate conformers may not exhibit the same stereoelectronic structural features present in the DFT-optimized ensembles, which would reduce the capability of the 3D GNNs to capture such effects. In cases where there were multiple MMFF94 conformers for a given molecule, we used up to 20 conformers as a form of data augmentation during model training, mapping each individual MMFF94 structure to the same descriptor of the DFT-optimized ensemble. This data augmentation strategy has been shown to slightly improve the modeling of conformer ensembles with little additional computation cost apart from conformer generation \citep{zhu2023learning}.

As the base 3D GNN architecture, we used DimeNet++ \citep{gasteiger2020fast}, which consistently performs on-par with or better than significantly more expensive equivariant 3D GNNs, especially for property prediction in relatively low-data tasks. To make use of atom features, we replaced DimeNet++’s default atom featurizer with the same one-hot encoded features employed in the 2D GNNs of \citet{haas2024rapid}, but bonds were not featurized. Furthermore, when applying each 3D GNN, we averaged the model’s predictions over <20 MMFF94 conformers to further reduce any noise relating to the choice of input conformer at test-time. Otherwise, the output of the 3D GNNs and the associated training protocols were analogous to those of the 2D GNNs.

\begin{figure}[t]
  \begin{center}
    \includegraphics[width=0.85\textwidth]{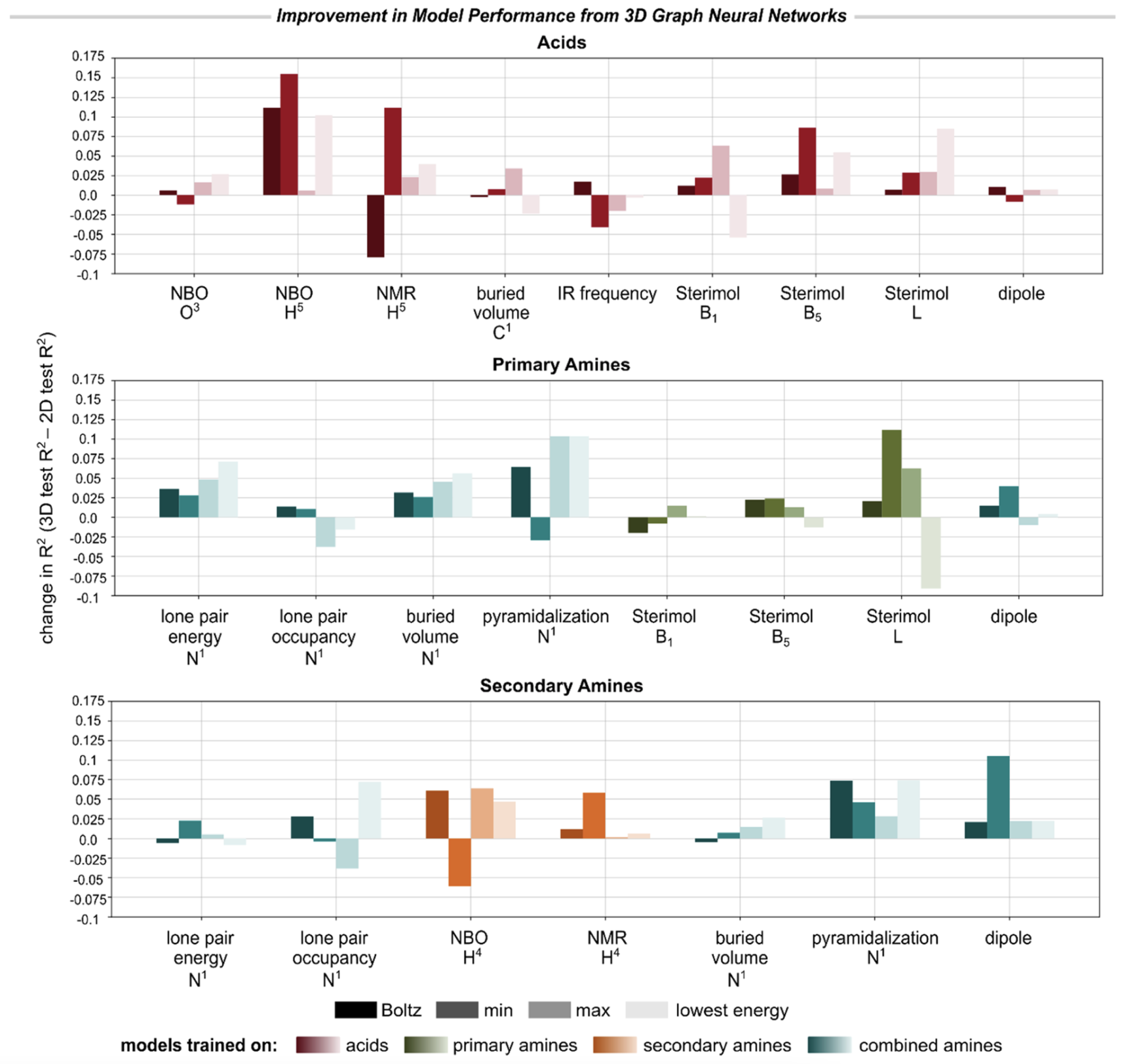}
  \end{center}
  \caption{Improvement in model performance on select molecule-, atom-, and bond-level descriptors when using DimeNet++ (operating on random MMFF94 conformers) versus the 2D GNN developed by \citet{haas2024rapid}, reported as the change in R2 on the original test sets in \citet{haas2024rapid}. }
  \label{fig:2d_vs_3D_gnns_haas}
\vspace{-10pt}
\end{figure}

Figure \ref{fig:2d_vs_3D_gnns_haas} reports the improvement in each 3D model’s performance (R2) compared to the corresponding 2D GNN when evaluated on the same test set.\footnote{A number of compounds could not be processed into RDKit conformers, and hence the 3D GNNs were trained and evaluated on marginally smaller (train/val/test) datasets than the 2D GNNs (acids: 7290/478/476; primary amines: 3168/495/494; secondary amines: 2425/500/498). Figure \ref{fig:2d_vs_3D_gnns_haas} compares the models when evaluated on the exact same test data for which conformers could be generated.}
Due to the greater cost associated with training the 3D models, we focused on modeling only a subset of descriptors for which the 2D GNNs had relatively poor accuracy, especially properties exhibiting sensitivity to stereoelectronic or geometric effects. In general, we found that the 3D GNN -- even when using MMFF94-level conformers -- led to slight or significant improvement in model accuracy for the majority of the modeled descriptors. The most significant improvements were observed for hydrogen atom NBO and NMR descriptors, Sterimol values, and amine pyramidalization. Only marginal improvement or diminished performance was observed for buried volumes, dipole moments, lone pair descriptors for the amines, and IR frequencies of the acids. 

Moreover, with the exception of IR frequencies, the 3D GNNs nearly always achieved lower MAE, even if the R2 did not improve (or was worsened), suggesting that encoding 3D molecular geometries improves the accuracy of descriptor prediction on average but may lead to greater sensitivity to outliers.
It is worth noting that DFT-derived descriptors have some inherent noise associated with them due to various sources of stochasticity during structure generation, geometry optimization, and conformer clustering. For geometry-sensitive properties like Sterimol values, error from the descriptors will propagate into the model, negatively impacting predictive accuracy. This noise may especially impact the descriptors of the lowest-energy conformers, which generally saw significantly reduced predictive accuracy for both the 2D and 3D models. The imperfect ability of the 3D GNNs to account for instances of stereoelectronic effects may also be due to the use of surrogate MMFF94 conformers as model inputs, or simply due to the rarity of these interactions in the training data.

\subsubsection{Overview of 3D GNN models}

DimeNet++ was employed as the backbone architecture of the 3D GNNs predicting atom-, bond-, or molecule-level descriptors of the conformer ensembles, but minor modifications were applied
to the atom embedding and readout layers of the original network, as described below. 

As a standard E(3)-invariant 3D graph neural network, DimeNet++ encodes 3D molecular structures, 
where each structure with $N$ atoms is represented as an attributed point cloud 
$P = \{(\mathbf{r}_i, \mathbf{x}_i)\}_{i=1}^N$ where $\mathbf{r}_i \in \mathbb{R}^3$ are the Euclidean coordinates of atom $i$ with features $\mathbf{x}_i \in \mathbb{R}^{d_x}$.

For any molecule-, bond-, or atom-level descriptor \textit{not} involving a hydrogen atom, we only included non-hydrogen atoms in the point cloud. For atom-level descriptors of hydrogen atoms (e.g., H\textsuperscript{4} NBO charge of secondary amines), we additionally include the hydrogen atoms of the acid or amine chemical moiety. Whereas the original implementation of DimeNet++ uses only one-hot representations of the element type as $\mathbf{x}_i$, we used additional (one hot) atom features computed with RDKit, including the element type, number of radical electrons, tetrahedral chirality, formal charge, aromaticity, ring size, degree, and total number of bonded hydrogens.

Given a structure $P$, DimeNet++ uses geometry-informed message passing to encode 
$\{ \mathbf{h}_i \}_{i=1}^N = f_{\text{DimeNet++}}(P)$, where 
$\mathbf{h}_i \in \mathbb{R}^{d_h}$ is the atom-specific representation of atom $i$. 
To adapt DimeNet++ to the descriptor prediction tasks, the models pool $\{\mathbf{h}_i\}_{i=1}^N$ into atom-, bond-, 
or molecule-level representations prior to predicting the scalar regression target $y$.

\[
\mathbf{h}_{\text{agg}} = \sum_{i=1}^N \mathbf{h}_i
\]

For atom-level descriptors:

\[
\mathbf{h}_{\text{atom}} = (\mathbf{h}_a, \mathbf{h}_{\text{agg}})
\]

\[
\hat{y}  = \text{MLP}_{\text{atom}}(\mathbf{h}_{\text{atom}})
\]

For bond-level descriptors:

\[
\mathbf{h}_{\text{perm}} = \text{MLP}_{\text{perm}}(\mathbf{h}_a, \mathbf{h}_b) + \text{MLP}_{\text{perm}}(\mathbf{h}_b, \mathbf{h}_a)
\]

\[
\mathbf{h}_{\text{bond}} = (\mathbf{h}_{\text{perm}}, \mathbf{h}_{\text{agg}})
\]

\[
\hat{y}  = \text{MLP}_{\text{bond}}(\mathbf{h}_{\text{bond}})
\]

For molecule-level descriptors:

\[
\mathbf{h}_{\text{mol}} = \mathbf{h}_{\text{agg}}
\]

\[
\hat{y} = \text{MLP}_{\text{mol}}(\mathbf{h}_{\text{mol}})
\]

where $(\cdot, \cdot)$ denotes concatenation in the feature dimensions. Here, $\text{MLP}_{\text{atom}}$, $\text{MLP}_{\text{bond}}$, and $\text{MLP}_{\text{mol}}$ are multi-layer perceptrons 
that map the pooled representations to scalar outputs. $\mathbf{h}_{\text{agg}}$ is an aggregated (sum-pooled) representation 
of the global molecule structure. $\mathbf{h}_a \in \{\mathbf{h}_i\}_{i=1}^N$ and/or 
$\mathbf{h}_b \in \{\mathbf{h}_i\}_{i=1}^N$ are representations of user-specified atoms in the molecule; in the case of predicting bond-level descriptors, atoms $a$ and $b$ form a covalent bond. $\text{MLP}_{\text{perm}}$ is a multi-layer perceptron that encodes a pair of (covalently-bonded) atoms into a permutation-invariant representation.

\subsubsection{Surrogate conformer generation with ETKDG/MMFF94}

The following workflow was employed to generate the surrogate conformers that were provided as input
to the 3D GNNs during training and inference. Given a SMILES, up to 100 conformers were embedded
with ETKDG in RDKit using an RMSD distance threshold of 0.25 \AA. Each individual conformer then
underwent up to 200 steps of MMFF94 optimization in RDKit. The optimized conformer ensembles were
then clustered with Butina clustering, using an initial threshold of 0.20 \AA. If there were more than 20
conformers remaining after clustering, the ensemble was iteratively re-clustered using increasingly 
RMSD thresholds (increments of 0.1 \AA) until the total number of clusters was less than or equal to 20.
The cluster centroids were then selected to form the final ensembles.

\subsubsection{3D GNN Training Protocols}
Each DimeNet++ model was trained with a mean squared error (MSE) loss until convergence, or for a
maximum of 3000 epochs. The best model over the course of training was selected for downstream
evaluation according to the mean absolute error (MAE) on the validation set. The Adam
optimizer was used in Pytorch with its default parameters and a constant learning rate.

For molecules that have multiple conformers in their surrogate MMFF94-optimized ensembles, multiple conformers were used as a form of data augmentation during training, with each conformer instance being mapped
to the same regression target. At the beginning of training, $n_c$ total
conformers were sampled (without replacement) from each molecule’s conformer ensemble. If the ensemble had fewer than $n_c$ conformers, then after sampling all possible conformers, the conformers were replaced and sampling was repeated until $n_c$ total conformers were obtained. This ensured that flexible molecules with many conformers were not
overrepresented during training. After sampling, the same $n_c$ conformers were used for each epoch. $n_c$ = 20 was used for the amines. $n_c=10$ was used for the acids due to the larger size of the acid datasets.

}{ \ \ }

\ifthenelse{\boolean{includeBackgroundOnHaas}}{
\clearpage
\subsection{Hyperparameters used for DimeNet++ in this work (and in \citet{haas2024rapid})}
\begin{table}[h!]
\centering
\caption{Default hyperparameters used for all DimeNet++ models.}
\begin{tabular}{|l|c|}
\hline
\textbf{Hyperparameter} & \textbf{Default value} \\ \hline
Learning rate & 0.0001 \\ \hline
Batch size & 128 \\ \hline
Size of hidden embeddings & 128 \\ \hline
Number of DimeNet++ blocks & 4 \\ \hline
Size of initial embedding & 64 \\ \hline
Size of basis embedding & 8 \\ \hline
Size of output embedding & 256 \\ \hline
Number of spherical harmonics basis functions & 7 \\ \hline
Number of radial basis functions & 6 \\ \hline
Cutoff distance for defining edges & 5.0 \AA \\ \hline
Maximum number of neighbors (edges per node) & 32 \\ \hline
Envelope exponent for smooth cutoff & 5 \\ \hline
Number of residual layers before skip connection & 1 \\ \hline
Number of residual layers after skip connection & 2 \\ \hline
Number of linear layers in DimeNet++ output blocks & 3 \\ \hline
Number of linear layers in $MLP_\text{mol}$, $MLP_\text{atom}$, $MLP_\text{bond}$ / $f_{\text{bond}}$ & 3 \\ \hline
Number of linear layers in $MLP_\text{perm}$ / $f_{\text{perm}}$ & 2 \\ \hline
Number of linear layers in $f_{\text{gate}}$ (this work only) & 3 \\ \hline
Activation function & swish \\ \hline
\end{tabular}
\end{table}
}{
\subsection{Hyperparameters used for DimeNet++ in this work}
\begin{table}[h!]
\centering
\caption{Default hyperparameters used for all DimeNet++ models.}
\begin{tabular}{|l|c|}
\hline
\textbf{Hyperparameter} & \textbf{Default value} \\ \hline
Learning rate & 0.0001 \\ \hline
Batch size & 128 \\ \hline
Size of hidden embeddings & 128 \\ \hline
Number of DimeNet++ blocks & 4 \\ \hline
Size of initial embedding & 64 \\ \hline
Size of basis embedding & 8 \\ \hline
Size of output embedding & 256 \\ \hline
Number of spherical harmonics basis functions & 7 \\ \hline
Number of radial basis functions & 6 \\ \hline
Cutoff distance for defining edges & 5.0 \AA \\ \hline
Maximum number of neighbors (edges per node) & 32 \\ \hline
Envelope exponent for smooth cutoff & 5 \\ \hline
Number of residual layers before skip connection & 1 \\ \hline
Number of residual layers after skip connection & 2 \\ \hline
Number of linear layers in DimeNet++ output blocks & 3 \\ \hline
Number of linear layers in $f_{\text{bond}}$ & 3 \\ \hline
Number of linear layers in $f_{\text{perm}}$ & 2 \\ \hline
Number of linear layers in $f_{\text{gate}}$ & 3 \\ \hline
Activation function & swish \\ \hline
\end{tabular}
\end{table}

}

\subsection{Categorization of trained models}
\label{app:tabulatedmodels}

\begin{table}[h!]
\centering
\caption{Overview of all models considered in this work. $n_c$: number of conformers included simultaneously in the model input.}
\begin{tabular}{@{}llll@{}}
\toprule
\textbf{\#} & \multicolumn{3}{c}{\textbf{Models Encoding Single Conformers}} \\ \midrule
\textbf{} & \textbf{active or random conformer?} & \textbf{optimization level} & $\mathbf{n_c}$ \\ \midrule
1          & active                               & DFT                         & 1              \\
2          & active                               & GFN2-xTB                      & 1              \\
3          & active                               & MMFF94                         & 1              \\
4          & random                               & DFT                         & 1              \\
5          & random                               & GFN2-xTB                      & 1              \\
6          & random                               & MMFF94                         & 1              \\
7          & random                               & DFT                         & 10             \\
8          & random                               & GFN2-xTB                      & 10             \\
9          & random                               & MMFF94                         & 10             \\ \midrule
\textbf{\#} & \multicolumn{3}{c}{\textbf{Models Encoding Sets of (Random) Conformers}} \\ \midrule
\textbf{} & \textbf{active included in set?}     & \textbf{optimization level} & $\mathbf{n_c}$ \\ \midrule
10         & no                                   & GFN2-xTB                      & 10             \\
11         & no                                   & MMFF94                         & 10             \\
12         & yes                                  & MMFF94                      & 10             \\
13         & yes                                  & GFN2-xTB                         & 10             \\
14         & yes                                  & DFT                         & 10             \\ \bottomrule
\end{tabular}
\label{tab:model_categorization}
\end{table}

\subsection{Additional results}
\label{app:additional_results}

\begin{table}[h!]
\centering
\caption{Absolute prediction performance for the 11 models showed in Figure \ref{fig:fidelity_comparisons}. Performance is reported by the mean absolute error on the test set, averaged across the three test sets. Model numbers correspond to the model numbers in Table \ref{tab:model_categorization}.}
\small
\begin{tabular}{@{}lccccccccccc@{}}
\toprule
\textbf{Target} & \textbf{\#1} & \textbf{\#2} & \textbf{\#3} & \textbf{\#4} & \textbf{\#5} & \textbf{\#6} & \textbf{\#7} & \textbf{\#8} & \textbf{\#9} & \textbf{\#10} & \textbf{\#11} \\ \midrule
B5 (max)          & 0.155            & 0.172            & 0.185            & 0.209            & 0.229            & 0.226            & 0.208            & 0.207            & 0.207            & 0.223             & 0.220             \\
B5 (min)          & 0.204            & 0.221            & 0.252            & 0.271            & 0.301            & 0.297            & 0.268            & 0.276            & 0.277            & 0.279             & 0.286             \\
L (max)           & 0.185            & 0.219            & 0.247            & 0.276            & 0.316            & 0.322            & 0.278            & 0.293            & 0.295            & 0.303             & 0.310             \\
L (min)           & 0.134            & 0.156            & 0.180            & 0.204            & 0.214            & 0.216            & 0.206            & 0.203            & 0.202            & 0.208             & 0.217             \\ \bottomrule
\end{tabular}
\end{table}

\end{document}